\pdfoutput=1

\documentclass[11pt]{article}

\usepackage{acl}

\usepackage{times}
\usepackage{latexsym}

\usepackage[T1]{fontenc}

\usepackage[utf8]{inputenc}

\usepackage{microtype}

\usepackage{inconsolata}

\newcommand{\minisection}[1]{\noindent{\bf #1}\hspace{0.6em}}

\usepackage{adjustbox}
\usepackage{graphicx}
\usepackage{enumitem}
\usepackage{url}
\usepackage{booktabs}
\usepackage{makecell}

\usepackage{amsmath}
\usepackage{multirow}

\title{\ours: An Improved Dataset for Entity Tracking in Texts}

\author{Li Zhang$^{1}$\thanks{\hspace{1.5mm}Work done as an intern at AI2.} \quad
  Hainiu Xu$^1$ \quad
  Abhinav Kommula$^{3}$ \\
  \textbf{Chris Callison-Burch}$^{1}$ \quad
  \textbf{Niket Tandon}$^2$ \\
  $^1$University of Pennsylvania\quad\quad $^2$Allen Institute for Artificial Intelligence \\
  $^3$University of California, Berkeley \\
  {\tt \{zharry\}@upenn.edu} \quad \tt{\{nikett\}@allenai.org}
}

\usepackage{tikz}
\usepackage{pgfplots}
\usepackage{xspace}

\newcommand{\theirs}{\textsc{OpenPI}\xspace}
\newcommand{\ours}{\textsc{OpenPI2.0}\xspace}

\usepackage{soul}

\newcommand{\squishlist}{
  \begin{list}{$\bullet$}
    { \setlength{\itemsep}{0pt}      \setlength{\parsep}{3pt}
      \setlength{\topsep}{3pt}       \setlength{\partopsep}{0pt}
      \setlength{\leftmargin}{1.5em} \setlength{\labelwidth}{1em}
      \setlength{\labelsep}{0.5em} } }
\newcommand{\reallysquishlist}{
  \begin{list}{$\bullet$}
    { \setlength{\itemsep}{0pt}    \setlength{\parsep}{0pt}
      \setlength{\topsep}{0pt}     \setlength{\partopsep}{0pt}
      \setlength{\leftmargin}{0.2em} \setlength{\labelwidth}{0.2em}
      \setlength{\labelsep}{0.2em} } }

 \newcommand{\squishend}{
     \end{list} 
 }

\renewcommand{\cite}{\citep}

\definecolor{lightgray}{gray}{0.9}
\definecolor{Box1Color}{RGB}{227, 236, 246}
\definecolor{Box2Color}{RGB}{248, 220, 225}
\definecolor{Box3Color}{RGB}{255, 238, 224}
\definecolor{cbBlue}{RGB}{0, 114, 178}
\definecolor{cbOrange}{RGB}{240, 228, 66}
\definecolor{cbGreen}{RGB}{0, 158, 115}
\definecolor{cbRed}{RGB}{213, 94, 0}
\definecolor{cbPurple}{RGB}{204, 121, 167}
\definecolor{cbSkyBlue}{RGB}{86, 180, 233}
\definecolor{cbGray}{RGB}{128, 128, 128}
\definecolor{CBF1}{RGB}{255,99,132}  
\definecolor{CBF2}{RGB}{54,162,235}  
\definecolor{CBF3}{RGB}{255,206,86}  
\definecolor{CBF4}{RGB}{75,192,192}  
\definecolor{CBF5}{RGB}{153,102,255} 
\definecolor{CBF1b}{RGB}{205,89,112}  
\definecolor{CBF2b}{RGB}{44,142,215}  
\definecolor{CBF5b}{RGB}{133,92,225}  

\begin{document}
\maketitle
\begin{abstract}
Much text describes a changing world (e.g., procedures, stories, newswires), and understanding them requires tracking how entities change. An earlier dataset, \theirs, provided crowdsourced annotations of entity state changes in text. However, a major limitation was that those annotations were free-form and did not identify salient changes, hampering model evaluation. To overcome these limitations, we present an improved dataset, \ours, where entities and attributes are fully canonicalized and additional entity salience annotations are added. On our fairer evaluation setting, we find that current state-of-the-art language models are far from competent. We also show that using state changes of salient entities as a chain-of-thought prompt, downstream performance is improved on tasks such as question answering and classical planning, outperforming the setting involving all related entities indiscriminately. We offer \ours for the continued development of models that can understand the dynamics of entities in text.\footnote{Our resources can be found at \url{https://github.com/allenai/openpi-dataset/tree/main/v2.0}.}
\end{abstract}

\section{Introduction}

Tracking entity states in procedural texts \cite{weston2015towards,DBLP:conf/iclr/BosselutLHEFC18,dalvi-etal-2018-tracking} is closely related to many NLP reasoning tasks. To name a few, question answering about events (e.g., \textit{should one use gloves when retrieving the tray from the oven}) often require knowledge of entity states (e.g., \textit{the tray becomes very hot while in the oven; gloves insulate heat}) \cite{tandon-etal-2019-wiqa,spiliopoulou-etal-2022-events,zhang-etal-2023-causal}; planning \cite{wang2022scienceworld,brohan2023can} largely involves actions upon entities resulting in state changes. While most recent work has relied on end-to-end language models (LMs) \cite{huang2022language}, recent work has shown that explicit modeling entity states benefits LMs in such tasks \cite{zhang-etal-2023-causal}. Procedural entity tracking is challenging in itself, requiring much understanding of an implicit environment as well as external knowledge of events and entities.


\begin{figure}[t!]
    \centering
    \includegraphics[scale=0.7]{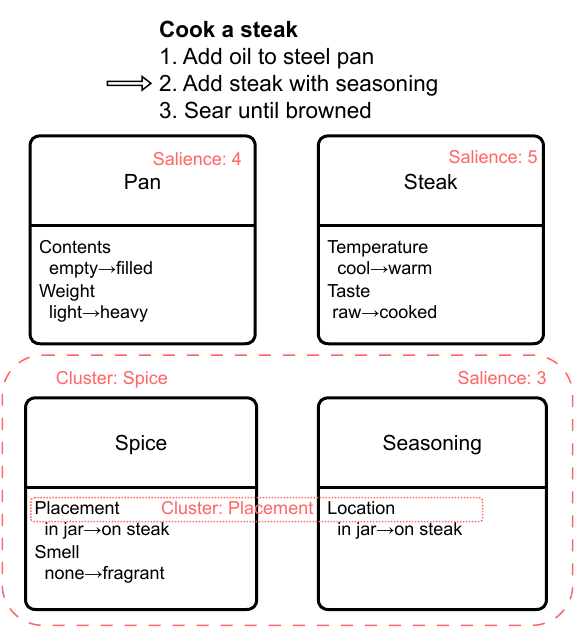}
    \caption{For each step in a procedure, \theirs annotates the state change of attributes of entities. Our \ours additionally (shown in red boxes and texts) canonicalizes the entities and attributes and includes their salience scores. }
    \label{fig:openpi1v2}
\end{figure}

We propose the \ours dataset which builds on \theirs (Open Procedural Inference) \cite{tandon-etal-2020-dataset}, a large-scale dataset for tracking entity states in procedural texts. \theirs contains annotations of entities, attributes, and state changes for each step (e.g., after the step ``set the pan in a heated oven'', the \textit{pan}'s \textit{temperature} was \textit{cool} before and \textit{hot} afterwards). \ours features two critical improvements (see Figure~\ref{fig:openpi1v2} for a demonstration of key features of \theirs and \ours):
\begin{enumerate}[topsep=0pt,itemsep=-1ex,partopsep=1ex,parsep=1ex,leftmargin=*]
    \item \textbf{Canonicalization}. Originally, different mentions of the same entity or attribute render evaluation difficult. Here, we prompt LMs to effectively cluster the entities and attributes.
    \item \textbf{Entity Salience}. Originally, a large amount of entities that undergo changes are listed in parallel. Here, we provide both human and model-predicted annotations of their salience.
\end{enumerate}
Regarding canonicalization, clustering different mentions (e.g., coffee maker, espresso machine) of the same entity allows for fairer evaluation. Moreover, as our task of predicting entities, attributes, and states is a generation task with imperfect and incomplete ground-truth references, we show that expanding each entity or attribute cluster with possible paraphrases (thus providing more references) is effective for reducing the false-negative rate. We then comprehensively report various state-of-the-art LMs' performance of entity tracking on \ours. 

Regarding entity salience, we provide both manually annotated and automatically predicted labels. We evaluate them based on correlation with ground-truth data, and show that LMs can reliably predict entity salience with a close-to-human performance. We argue that salient entities act as a means of compression of the most critical information in procedural texts, similar to saliency maps in computer vision \cite{simonyan2013deep}. We proceed to qualitatively and quantitatively show that salient entities, as chain-of-though of LM prompting, benefit downstream tasks such as question answering and classical planning, while reducing cost by excluding less important entities in the prompt.

\ours have following advantages:
\begin{enumerate}[topsep=0pt,itemsep=-1ex,partopsep=1ex,parsep=1ex]
    \item The canonicalization of entities and attributes (\S\ref{sec:canonicalization}) that facilitates evaluation (\S\ref{sec:eval_entity_tracking});
    \item The salience of entities (\S\ref{sec:salience}) that improves performance on downstream tasks (\S\ref{sec:downstream}).
\end{enumerate}

\begin{figure}[t!]
    \centering
    \includegraphics[width=\columnwidth]{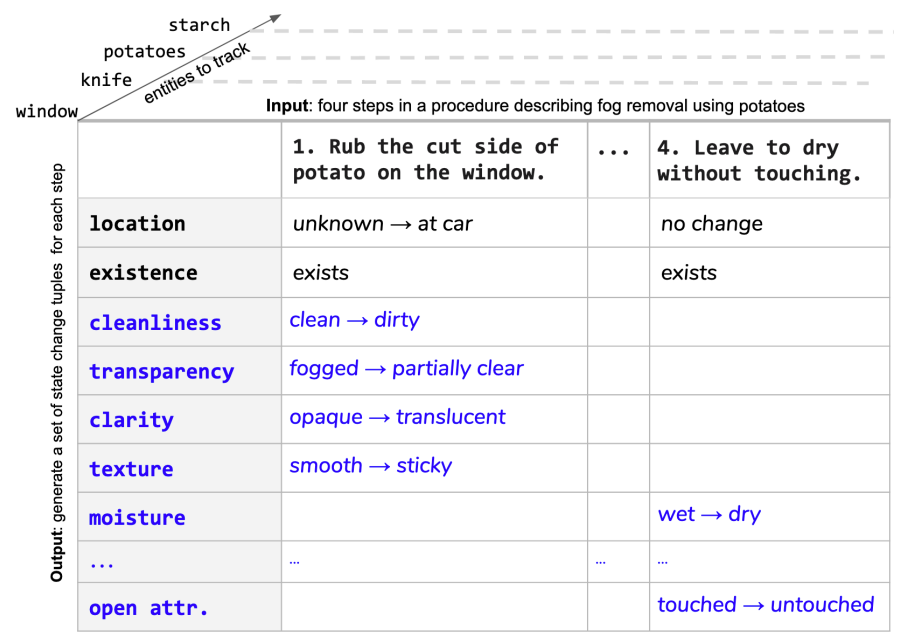}
    \caption{An example from the original \theirs dataset  \cite{tandon-etal-2020-dataset}.}
    \label{fig:orig_openpi}
\end{figure}

\section{The Original \theirs Dataset}
\label{sec:original}

Our work \ours builds upon the \theirs dataset \cite{tandon-etal-2020-dataset} that tracks entity state changes in procedural texts with an open vocabulary. The procedures are extracted from wikiHow, a web resource containing instructions  of everyday tasks. As exemplified in Figure~\ref{fig:orig_openpi}, the input is a procedure which includes a goal (e.g., ``remove fog using a potato'') and a sequence of ordered steps (e.g., ``rub the cut side of potato on the window''). For each step, the output is an array of 4-tuples describing an entity state change. Each 4-tuple contains an entity, an attribute, a state before the step, and a state after the step (e.g., \textit{the window}'s \textit{texture} was \textit{smooth} before and \textit{sticky} after). The task is thus equivalent to predicting an entity state matrix given a procedure, where the axes are step, entity, and attribute, while the value is the before and after states. The data is annotated via crowdsourcing and manually validated. 

However, \theirs lacks canonicalization of entities and the differentiation of salient entities. In our work of \ours, we will address both issues using state-of-the-art models.

\begin{table*}
\small
\centering
    \begin{adjustbox}{max width=\textwidth}
    \begin{tabular}{ll}
        \toprule
        Role & Content \\
        \midrule
        User & \makecell[l]{I am trying to make coffee. First, I put some coffee beans and tap water into the corresponding compartment \\
        of the espresso machine. Then, I select the desired type of coffee to make produced. Then I put a mug under \\
        the espresso machine and press start. Do you get it?} \\[4mm]
        Assistant & Yes. \\[1mm]
        User & \makecell[l]{
        We have the following objects: \textit{water}, \textit{coffee maker}, \textit{coffee machine}, \textit{mug}, \\\textit{espresso machine}. Group those that refer to the same thing. You must include all the provided entities. Do \\
        not add any entity that is not provided in the list.}  \\[4mm]
        Assistant & \makecell[l]{<start of generation> The grouped objects are:\\  - ['water']\\  - ['coffee maker', 'coffee machine', 'espresso machine']\\  - ['mug']} \\
        \bottomrule
    \end{tabular}
    \end{adjustbox}
    \caption{Our chosen prompt for entity and attribute clustering.}
\label{tab:canon_prompt}
\end{table*}

\section{Canonicalization}\label{sec:canonicalization}
In the original \theirs dataset, the entities and attributes that undergo change were written by crowd workers. Consequently, the dataset contains different ways of expressing the same entity (e.g., \textit{coffee maker}, \textit{coffee machine}, \textit{espresso machine} in a coffee-making procedure) or attribute (e.g., \textit{texture}, \textit{smoothness}, \textit{sheen} of a paint). Canonicalization by clustering the entities and procedures is thus important for two reasons: 1) it facilitates evaluation, especially in a generation setting, where a model might be wrongly penalized for predicting the paraphrase of some correct entity or attribute; 2) it facilitates further annotation of features such as salience (\S\ref{sec:salience}) of the entities and attributes. Here, we describe efforts to canonicalize the entities and attributes in the evaluation set of \theirs.

\subsection{Clustering Entities and Attributes} \label{sec:clustering}
While canonicalization seems straightforward, it is non-trivial in \ours because clustering is highly context-dependent. For example, the entity \textit{torso} and \textit{paper chunk} usually have nothing to do with each other, but in fact refer to the same thing in a procedure of ``making a paper bird.'' 

\minisection{Clustering} Due to the contextual nature of the task, we prompt one of the state-of-the-art LMs
\texttt{gpt-3.5-turbo} (a.k.a. ChatGPT)\footnote{\url{platform.openai.com/docs/models/gpt-3-5}} as shown in Table~\ref{tab:canon_prompt}. We use 3-shot prompting, meaning that the complete prompt includes three handwritten examples and the prompt header of the example to be inferred, only containing the ``User'' role. The temperature is as default (0.7) and so are other hyperparameters. We aggregate output from five runs of ChatGPT as the final entity cluster and three runs for attribute cluster, as doing so is found to be empirically superior than a single-pass generation.\footnote{With results from multiple runs, entity clusters are greedily selected based on their number of occurrences. For instance, if \texttt{(pan, cookware, container)} occurred four times whereas \texttt{(pan, pot)} just once, then the former will be added to the final cluster.}

To see if our model can cluster entities and attributes effectively, we evaluate the results using cluster-level precision, recall, and F1 scores with exact match against a set of manually-labeled clusters from 20 procedures in the development set. 


\begin{table}[t!]
    \small
    \centering
    \begin{tabular}{lll} \toprule
              & Entity & Attribute \\ \midrule
    Cluster Recall    & .425   & .881      \\
    Cluster Precision & .593   & .906      \\
    Cluster F1        & .495   & .893     \\\bottomrule
    \end{tabular}
    \caption{Evaluation of entity and attribute clustering.}
    \label{tab:cluster_performance}
\end{table}

We see that ChatGPT scores better in clustering attributes compared to entities. Error analysis shows that two factors contribute to this performance discrepancy. First, most attributes describe the physical properties of an entity. Therefore, attribute clusters are less context-dependent compared to entity clusters. Second, many attributes are shared amongst entities. For instance, out of 1,145 attribute annotations in the development set, 204 of them are \textit{"location"}.

\begin{table*}[]
\centering
\small
\begin{tabular}{lllll|llll|ll}
\toprule
                 & \multicolumn{4}{c|}{schemata (global)}                                  & \multicolumn{4}{c|}{schemata (local)} & \multicolumn{2}{c}{states} \\ \midrule
                 & F1        & F1 + exp       & BS               & BS + exp              & F1        & F1 + exp       & BS              & BS + exp & acc. & BS \\ \midrule
\texttt{gpt-3.5-turbo}    & .151 & .249 & .843 & .869 & .025 & .039  & .798 & .804 & .074 & .600 \\
\texttt{text-davinci-003} & .362 & .450 & .891 & .920  & .130 & .155  & .798 & .810 & .225 & .682 \\
\texttt{LLaMA 65B}        & .129 & .174 & .799 &  .820 & .045 & .060 & .801 & .800 &   .102  & .577 \\ 
\bottomrule
\end{tabular}
\caption{Exact match F1 or accuracy and BERTScore on the schemata and states prediction sub-tasks, with and without cluster expansion. The schemata sub-task is evaluated both globally (per-procedure) and locally (per-step).}
\label{tab:perf_1}
\end{table*}

\minisection{Cluster expansion} Though the existing entities and attributes are now clustered in \ours, there may still be other paraphrases that a model might rightfully predict and wrongfully penalized for. Thus, we again prompt ChatGPT to expand the clusters by generating paraphrases given a cluster of entities or attributes (prompt omitted). 

To evaluate the quality of entities and attributes generated from the expansion, we manually rate 20 procedures and find that 83.3\% of the generated, paraphrased entities and 59.4\% attributes are correct.
This is largely because entity names are oftentimes self-explanatory and less context-dependent whereas the attribute names and their meanings are highly dependant on the context. 

\begin{table}[t!]
    \centering
    \small
    \begin{tabular}{lllll} \toprule
    & \multicolumn{4}{c}{complete} \\ \midrule
      & F1        & F1+exp       & BS               & BS+exp \\ \midrule
    \texttt{gpt-3.5-turbo} & .016 & .016 & .772 & .790  \\
    \texttt{text-davinci-003} & .034 & .034 &  .807  & .821  \\
    \texttt{LLaMA 65B} & .117 & .117 & .429 & .440 \\ 
\bottomrule     
    \end{tabular}
    \caption{Exact match F1 and BERTScore of complete sentences including an entity, an attribute, a pre-state, and a post-state, following the original \theirs paper. Canonicalization and expansion lead to little help for exact match as it is only done on entity and attribute clusters, while the state names can still be expressed in many ways, causing false negatives.}
    \label{tab:perf_2}
\end{table}

\begin{table}[t!]
    \small
    \centering
    \begin{adjustbox}{max width=\columnwidth}
    \begin{tabular}{lllll}
    \toprule
    & Correct & No change & Nonsense & Missing \\ \midrule
    \texttt{003} & 585 (82.3\%) & 106 (15.0\%) & 	14 (2.0\%) & 383 (20.3\%) \\
    \texttt{3.5} & 303 (59.4\%) & 173 (33.9\%) & 34 (6.7\%) & 218 (42.7\%) \\ \bottomrule   
    \end{tabular}
    \end{adjustbox}
    \caption{Error analysis on the schemata prediction task of \texttt{text-davinci-003} and \texttt{gpt-3.5-turbo}.}
    \label{tab:error_analyis}
\end{table}

\subsection{Utility: Evaluation of Entity Tracking} \label{sec:eval_entity_tracking}

Just as the original evaluation set of \theirs, \ours is meant to benchmark models on entity tracking -- given a step in a procedure, predicting the state changes that entities and their attributes undergo. With the entities and attributes in \ours now fully canonicalized, evaluation can be done more fairly. To start with, we follow \citet{tandon-etal-2020-dataset} and have models predict one \textbf{complete} sentence: ``\textit{attribute} of \textit{entity} is \textit{pre-state} before and \textit{post-state} afterwards'', which is then compared to such sentences in the ground-truth data (Table~\ref{tab:perf_2}). We further make the evaluation more fine-grained by formulating two sub-tasks: i. predicting \textbf{schemata}, namely the entities and their corresponding attributes given a step (e.g., given ``turn on the oven'', the \textit{temperature} of the \textit{rack} undergo state changes), and ii. predicting the change of \textbf{states} given a step, an entity and an attribute (e.g., given the previous information, the state change is from \textit{cool} to \textit{hot}). This evaluation of first predicting a skeleton tensor of entities and attributes is highly practical, with a notable advantage over previous work (\S\ref{sec:related_work}) in closed-domain entity tracking, where states are predicted using \textit{given} entities and attributes.

On the development set, we run three state-of-the-art LMs: \texttt{gpt-3.5-turbo}, \texttt{text-davinci-003}\footnote{\url{platform.openai.com/docs/models/gpt-3-5}} \cite{brown2020language}, and the open-source LLaMA 65B \cite{touvron2023llama}. For each model, we start by separately tackling each of the two sub-tasks\footnote{To avoid error propagation, for states prediction, the ground-truth entities and attributes are provided.}; namely, a model first predicts attributes of entities (schemata) given a step, and then predicts a pre-state and a post state (states) given the gold entity-attribute pair.  All experiments are via 1-shot prompting. See details on prompt formulation in Appendix~\ref{sec:details_eval_entity_tracking}.

\begin{table*}
\small
\centering
    \begin{tabular}{ll}
        \toprule
        Role & Content \\
        \midrule
        User & \makecell[l]{Here are some instructions on making coffee.\\- Buy fresh coffee beans.\\- Grind the coffee beans.\\- ...\\Now, I will provide you with a series of objects, and you will assign scores on a scale of 1-5 to them based on\\their importance in the instruction. Your answer should strictly be a numerical score followed by a one-\\sentence explanation.}  \\
        Assistant & Sure, I can help you with that. Please provide the objects.\\
        User & Coffee bean \\
        Assistant & <generation> 5 - the coffee beans are the most important ingredient in making coffee. \\
        \bottomrule
    \end{tabular}
    \caption{Our chosen prompt for predicting global or procedure-wide entity salience. For local salience, the wording is similar with only one step provided.}
\label{tab:salience_prompt}
\end{table*}


For all settings, we consider both exact match (F1 for schemata and complete sentence prediction and accuracy for states prediction) and BERTScore \cite{bert-score} based on \texttt{deberta-xlarge-mnli} \cite{he2021deberta}. 

For the schemata prediction sub-task (Table~\ref{tab:perf_1}), the atomic unit to be evaluated is an entity-attribute pair. We consider both a \textit{global} evaluation, where predictions are made per-procedure (e.g., what attributes of entities undergo state changes in the procedure), and a \textit{local} evaluation, where predictions are made per-step. This categorization will reappear in \S\ref{sec:salience_eval}. Schemata prediction is naturally influenced by our entity and attribute clusters. Hence, for exact match we report F1 scores based on exact matches where any entity-attribute prediction that falls under an cluster, obtained by taking a Cartesian product of an entity cluster and an attribute cluster, is considered a true positive. For BERTScore, we calculate the maximum score of a prediction against all entity-attribute strings within all ground-truth clusters. Then, we report the mean score among all predictions as a macro average.

The states prediction sub-task (Table~\ref{tab:perf_1}) is much more straightforward as the entity-attribute pairs are provided and a model only needs to predict a pre-state and a post-state for each. Thus, we simply report the exact match accuracy and BERTScore for each state.

\subsection{Discussion and Error Analysis}

We observe that the predicting attributes of entities that undergo state changes is a highly challenging task even for state-of-the-art LMs. Although evidently, expansion of clusters improves performance (fairly, as we have shown that the generated paraphrases are mostly correct), false-negatives that result in underestimation of models cannot be eliminated entirely. One interesting observation is that \texttt{text-davinci-003} greatly outperforms the supposedly more superior \texttt{gpt-3.5-turbo}. To gain even more insights into models' behavior, we analyze the model output for the schemata prediction sub-task. For each step, we annotate each entity-attribute prediction based on three labels:

\begin{itemize}[topsep=0pt,itemsep=-1ex,partopsep=1ex,parsep=1ex]
    \item Correct, where the entity-attribute indeed go through some changes;
    \item Incorrect, because the entity-attribute actually does not go through any changes;
    \item Incorrect, because the entity-attribute is non-sensical.
\end{itemize}

In addition, we add any entity-attribute pairs that should have been predicted as going through some change, to measure models' recall. We randomly sample 20 procedures to perform this error analysis and the results are shown in Table~\ref{tab:error_analyis}.

Regarding precision, we find that while the majority of the predicted entities are correct, many of the predicted associated attributes are generic ones that do not undergo any change either locally or globally. For example, for the step ``Purchase a blackboard eraser'', the attributes predicted by \texttt{text-davinci-003} for the entity \textit{eraser} are \textit{location} (correct), \textit{cleanness} (no change locally), \textit{shape}, and \textit{size} (no change globally). The issue is much more pronounced with \texttt{gpt-3.5-turbo}, with predictions such as \textit{location} of \textit{seller}, \textit{name} of \textit{brand}, etc, despite that the prompt clearly explains the desired output with an example. We attribute such performance discrepancy to \texttt{gpt-3.5-turbo}'s decreased ability to follow examples and its inability to understand nuanced instructions. Regarding recall, both models fail to predict many attributes that the human annotator deems changing. Upon qualitative inspection, most of these missing attributes are no less salient than the predicted ones.

We leave to future work the resolution of these issues, which can be mitigated by re-prompting the models by validating if the predicted attributes indeed undergo changes, or simply have them predict the state changes altogether in the first place. 

\begin{table}
\small
\centering
    \begin{tabular}{lll}
        \toprule
         & \multicolumn{1}{c}{Annotations} & \multicolumn{1}{c}{Predictions} \\
         & Human (A2)  & LM  \\ \midrule
        Global & .759 & .719  \\
        Local & .578 & .400  \\
        \bottomrule
    \end{tabular}
    \caption{Pearson' $r$ between model prediction and human annotations (A1) of entity salience.}
\label{tab:correlation}
\end{table}

\section{Salience} \label{sec:salience}
The original \theirs is annotated with many parallel entities in each procedure. Often, they vary greatly by importance in accomplishing the task. For example, in a procedure of ``cooking a steak'', entities \textit{fish}, \textit{oven}, \textit{gloves}, and \textit{spice rack} might all be involved, while some are more indispensable than the rest. Intuitively, the knowledge of entity salience helps models focus on what matters in downstream tasks (\S\ref{sec:downstream}). In \ours, we define two types of entity salience: the global salience refers to the importance of an entity in accomplishing the goal of the procedure, whereas the local salience refers to that in a step. 

\begin{figure*}
    \centering
    \includegraphics[width=\textwidth]{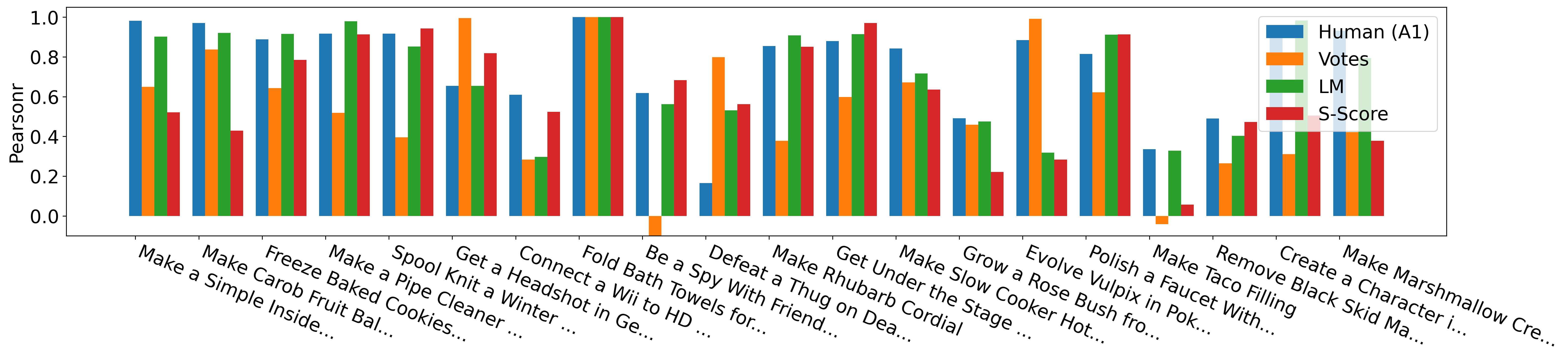}
    \caption{Per-procedure correlation of global entity salience between each set of annotations and the ground-truth human annotations.}
    \label{fig:salience_correlation}
\end{figure*}

\subsection{Annotations} \label{sec:salience_annotations}

\minisection{Human Labeling} To first procure ground-truth salience labels, two experts (referred to as A1 and A2) annotated entity salience in the first 20 procedures in the development set as the gold standard of entity salience. We devise and follow these annotation criteria in a Likert scale:
\begin{enumerate}[topsep=0pt,itemsep=-1ex,partopsep=1ex,parsep=1ex]
    \item [5:] without or without mentioning this entity, the procedure or step cannot be done at all (e.g., \textit{lemon} in “Wash faucet with lemon”)
    \item [4:] without this entity, another entity of the same type can be used as a replacement, perhaps with worse outcome or more efforts (e.g., pan in “Sear a salmon” - can also use \textit{grill})
    \item [3:] without this entity, the procedure or step can be done in principal, though with slightly worse outcome or more efforts (e.g., \textit{glove} in “Cut off tough branches of a bonsai plant”)
    \item [2:] without this entity, the procedure or step can be done, though with negligibly worse outcome or more efforts (e.g., \textit{vacuum cleaner} in “Drill holes in the wall”)
    \item [1:] the entity appears in the procedure or step rather gratuitously, and the lack thereof makes no difference
    \item [0:] the entity is irrelevant to the procedure or step
\end{enumerate}

Subjectivity is inevitable even though we strive to minimize subjectivity using this fine-grained scale to capture nuanced situations (e.g., an entity that frequently appears that can be easily replaced versus one that appears only once but is irreplaceable). In later sections, we will see how this scale leads to reasonable inter-annotator agreement and favorable performance on downstream tasks.



\minisection{LM Prediction} We prompt \texttt{gpt-3.5-turbo}, as before, to automatically predict salience. Table~\ref{tab:salience_prompt} shows an example prompt for predicting global salience. As before, we use the default hyperparameters with a temperature of 0.7. We parse the result by extracting the first digit from the generation as the score, and default to 1 whenever impossible. \vspace{+2mm}


\subsection{Evaluation} \label{sec:salience_eval}

To first holistically evaluate the modelling of salience, we report pairwise Pearson's correlation coefficients between each set of labels above and the annotations of human A1. In Table~\ref{tab:correlation}, we report a ``macro correlation'', namely the mean of correlation of salience scores in each procedure.\footnote{To avoid NaN due to constant input array, a 0 is appended to each array as smoothing.} First, the correlation between the two annotators is high but imperfect, implying  subjectivity in the annotation of entity salience. In comparison, the LM predictions come close with especially impressive predictions for global salience. 

To understand when and how entity salience can be subjective among humans, in Figure~\ref{fig:salience_correlation} we show salience correlation for the first 20 procedures. Some with low inter-human correlation such as ``Defeat a Thug'' expose a document-frequency problem: one human labels the entity \textit{you (actor)} with a salience of 5, believing that without the \textit{actor} the procedure cannot be undertaken at all, while the other labels 1, believing that the \textit{actor} occurs in every procedure and thus the mention of it makes no difference. Other procedures such as ``Make Taco Filling'' expose a cluster-individual problem: all individual ingredients are grouped under one cluster, for which annotators are divergent on the salience. Both issues give rise to subjectivity and can be mitigated by better defining entity salience in a potential future annotation process. 


\begin{table*}
\small
\centering
    \begin{adjustbox}{max width=\textwidth} 
    \begin{tabular}{ll}
        \toprule
        Role & Content \\
        \midrule
        User & \makecell[l]{A person walks into a dark room in a \textbf{freezing windy night}. In the very beginning before anything is done, \\how likely is it that she can see what's in the room? Answer on a scale of 1-5, with a 1-sentence explanation.}  \\[5pt]
        Assistant & 1 - it is highly unlikely for the person to see anything in a dark room without any source of light. \\[5pt]
        User & \makecell[l]{Next, she lights up a torch and hangs it near a window. At this point, how likely (same question)...} \\[5pt]
        Assistant & \makecell[l]{5 - hanging the torch near a window maximizes the distribution of light in the room, greatly enhancing the \\visibility and increasing the likelihood of seeing what's inside.} \\[5pt]
        \textsuperscript{1}User & \makecell[l]{Next, she \textbf{opens the window}. At this point, how likely (same question)...} \\[5pt]
        \textsuperscript{1}Assistant & \makecell[l]{\textbf{5} - opening the window allows \textbf{natural light} to enter the room, combining with the torchlight to provide \\ample illumination, greatly increasing the likelihood of seeing what's in the room.} \\[5pt] \midrule
        \textsuperscript{2}User & \makecell[l]{Next, she opens the window. The most salient entities for this step are the window, the wind, and the flames.\\How does each of these entities change?} \\[5pt]
        \textsuperscript{2}Assistant & \makecell[l]{When she opens the window, the following changes occur:\\Window: The \textbf{window is now open}, allowing for air and potentially some outside light to enter the room.\\Wind: The wind becomes more noticeable and \textbf{enters the room through the open window}, causing a draft\\and potentially affecting the flames of the torch.\\Flames: The flames of the torch may flicker or change direction due to the entry of wind through the open\\window, \textbf{altering their shape and intensity}.} \\[5pt]
        \textsuperscript{2}User & \makecell[l]{With such information, at this point, how likely is it that she can see what's in the room?} \\[5pt]
        \textsuperscript{2}Assistant & \makecell[l]{\textbf{3} - opening the window allows some outside light to enter, the presence of wind and potential \textbf{disruption to} \\\textbf{the flames} may still impede visibility to some extent.}
        \\ \bottomrule
    \end{tabular}
    \end{adjustbox}
    \caption{\texttt{gpt-3.5-turbo}'s performance on a CREPE-like example. The interactions with and without using entity salience are both shown. Critical information is illustrated in bold.}
\label{tab:crepelike_prompt}
\end{table*}

\subsection{Utility: Downstream Applications} \label{sec:downstream}

We argue that entity salience is an effective way to compress information expressed by procedural texts. In other words, states of the most salient entities are sufficient for downstream tasks where entity tracking can be applied to. We provide both qualitative and quantitative evidence on two datasets both in the domain of procedural texts.  

\subsubsection{Question Answering} CREPE \cite{zhang-etal-2023-causal} is a dataset for procedural question answering and causal reasoning. Given a procedure (e.g., steps of ``baking a cake''), a model predicts the change of likelihood of some event occurrence (e.g., ``there is a pleasant smell coming from the oven'') after the execution of each step (e.g., ``set the tray in the oven for 45 minutes'').

First, we show a qualitative example in Table~\ref{tab:crepelike_prompt} in the format of CREPE. Clearly, the model's third response is wrong, and the user's first utterance implies that the \textit{torch} would have been \textit{extinguished} by the cold wind or at least flickering, making it harder to see what's in the room. In contrast, the model falsely asserts that there would be \textit{natural light}, which is impossible given the procedure occurs at night. However, if we use the same prompt as \S\ref{sec:eval_entity_tracking} and Table~\ref{tab:salience_prompt} to first get access to the most locally salient entities, have the model predict their state changes, and use the such information as a chain-of-thought, the model is able to answer correctly. Specifically, the model now recognizes that the \textit{window} is \textit{open}, causing the wind to enter the room, in turn causing the \textit{flames} to \textit{flicker}. With this information equipped, the final predicted answer is now correct as the likelihood to see clearly in the room has decreased. Note that the step of opening the window also involves many other less salient entities, such as \textit{person}, \textit{hands}, \textit{windowsill}, \textit{smell}, etc., which are filtered out by predicted salience.

\begin{table}[t!]
    \centering
    \small
    \begin{tabular}{llll}
    \toprule
                              & dev  & test & num. ents per step \\ \midrule
    ChatGPT                   & .348 & .362 & - \\
    +all ents                 & .392 & .369  & 2.9 \\
    +ents sal\textgreater{}=5 & \textbf{.402} & \textbf{.370} & 1.8      \\
    \bottomrule
    \end{tabular}
    \caption{Model performance on CREPE, with and without including state changes of salient entities.}
    \label{tab:crepe_perf}
\end{table}

Following the qualitative study, we then quantitatively evaluate a \texttt{gpt-3.5-turbo} model on CREPE in a zero-shot manner. As the labels in CREPE are ternary (an event is more, less, or equally likely to happen after a step), we directly prompt the model to provide one of these labels instead of giving a likelihood score as in Table~\ref{tab:crepelike_prompt}, while keeping the rest of the designs the same. 
We compare three settings, in which for each step given an event, the model:
\begin{enumerate}[topsep=0pt,itemsep=-1ex,partopsep=1ex,parsep=1ex,leftmargin=*]
    \item makes a prediction directly;
    \item is provided with all entity state changes (based on \S\ref{sec:eval_entity_tracking}), before making a prediction;
    \item is provided with state changes of entities with a predicted salience score (based on \S\ref{sec:salience_annotations}) no less than a threshold, before making a prediction.
\end{enumerate}

Table~\ref{tab:crepe_perf} clearly shows that provided entity state changes as chain-of-though in the prompt improves the performance. Moreover, excluding information of less salient entities does not decrease, and even in some cases improve performance. By only keeping the entities that have a predicted salience score of 5, which constitute about half of all entities, the prompt becomes shorter and thus reducing cost by about 14\% for the entire experiment, while achieving slightly higher performance.

\subsubsection{Formal Planning based on Texts} A strand of work has focused on using LMs to plan in a structured and symbolic manner \cite{liu2023llm+, xie2023translating}. Specifically, the task is to convert a a textual description of a procedure to a planning domain definition (PDDL), which can be deterministically solved to find a plan. An instance of PDDL consists of a domain file, which models the pre-conditions and effects of permitted actions, and a problem file, which models the entity states initially and eventually. Here, we show a qualitative example of how identification of salient entities can improve the quality of a generated domain file that models actions. 

\begin{table}[t!]
\small
\centering
\begin{adjustbox}{max width=\columnwidth}

    \begin{tabular}{l}
        \toprule
        User \\
        \midrule
        \makecell[l]{Provide a PDDL domain file representing a procedure\\"Survive in rainy woods":
\\1. Find natural shelter using the environment
\\2. Set up containers in the open
\\3. Stay dry during a rain
\\4. Stay hydrated
\\After that, also provide a problem file of "collect water\\to drink".}  \\\midrule
        Assistant \\
        \midrule
        \makecell[l]{Here's a domain file:
\\(define (domain survive-in-rainy-woods)
\\\quad requirements, types, predicates are omitted
\\\quad(:action find-shelter
\\\quad\quad:parameters (?shelter - shelter ?loc - location)
\\\quad\quad:precondition (at ?shelter ?loc)
\\\quad\quad:effect (and (sheltered ?shelter)))
\\\quad(:action set-up-container
\\\quad\quad:parameters (?container - container ?loc - location)
\\\quad\quad:precondition (and (at ?container ?loc) (sheltered\\?container))
\\\quad\quad:effect (and (contains \textbf{?container ?water})))
\\\quad(:action stay-dry-during-rain
\\\quad\quad:parameters (?obj - object ?loc - location)
\\\quad\quad:precondition (and (at ?obj ?loc) (raining ?loc))
\\\quad\quad:effect (and (not (raining ?loc)) (dry ?obj)))
\\\quad(:action stay-hydrated
\\\quad\quad:parameters (?obj - object ?container - container)
\\\quad\quad:precondition (and (at ?obj ?container) (contains\\\quad\quad\textbf{?container ?water}))
\\\quad\quad:effect (and (not (contains ?container ?water))\\\quad\quad(\textbf{hydrated ?obj}))))
\\Here's a problem file:
\\(define (problem collect-water)
\\\quad objects are omitted
\\\quad(:init
\\\quad\quad(at water-bottle woods)
\\\quad\quad(at shelter woods)
\\\quad\quad(at player woods)
\\\quad\quad(raining woods)
\\\quad\quad(collecting-water woods))
\\\quad(:goal (\textbf{hydrated player})))
}  \\[5pt]
        \bottomrule
    \end{tabular}
    \end{adjustbox}
    \caption{\texttt{gpt-3.5-turbo}'s performance on a wikiHow-to-PDDL-like example. The interactions without using entity salience are both shown. Critical information is illustrated in bold.}
\label{tab:pddllike_prompt}
\end{table}

Table~\ref{tab:pddllike_prompt} shows an example where an LM provides a domain file based on the procedure ``Survive in the woods'' and a problem file based on the task ``collect water to drink''. Clearly, the goal configuration \texttt{(hydrated player)} can only be realized via the action \texttt{stay-hydrated}, whose pre-condition \texttt{(contains ?container ?water)} can only be satisfied in the second step. However, the LM fails to recognize that the third step ``Stay dry during a rain'' has the byproduct effect of filling up the containers set up in the second step. In this case, we might simply first provide the LM the salient entities for each step, and the LM successfully fixes the third action \texttt{stay-dry-during-rain} so that it has the effect of containing containing water. Therefore, the problem file can now be solved reasonably with a sequence of all four actions. We leave to future work a larger-scale experiment of the application of salient entity states to planning.


\section{Resulting Dataset: \ours} \label{sec:dataset}
By adding canonicalization of entities and attributes as well as salience of entities to the evaluation set of the \theirs dataset, we now fully present \ours. As the procedures and entity state annotations have not changed, \ours still has 55 procedures with 5.0 steps on average. These procedures are collected from wikiHow and their topics are everyday activities. \ours also inherits the original entity-attribute-state changes annotated by crowd workers. After canonicalization, there are 356 canon entities each with 7.6 unique mentions and 5.5 expanded mentions on average, 3240 canon attributes, each with 3.0 unique mentions and 3.3 expanded mentions on average, and 1193 before-after states in the development set. The quality of clustering and expansion and be evidenced in \S\ref{sec:clustering}. Regarding salience labels (on a scale of 1 to 5), the global salience of entities has a mean of 3.5 and standard deviation of 1.4; the local salience of entities has a mean of 3.4 and standard deviation of 1.5.


\section{Related Work} \label{sec:related_work}

\paragraph{Entity State Tracking}

Prior work on entity state tracking spans various disciplines of AI. For instance, object tracking, a sub-task of entity state tracking, has led to much work in both robotics \cite{wang2007simultaneous} and computer vision \cite{comaniciu2003kernel}. In NLP, early efforts focus on synthetic, closed-domain data \cite{weston2015towards, long-etal-2016-simpler} and more recent ones shift attention to real-world procedures \cite{bosselut2017simulating, dalvi-etal-2018-tracking,gupta-durrett-2019-tracking,du-etal-2019-consistent,mysore-etal-2019-materials} with a closed set of entities and attributes. The only open-ended dataset to our knowledge is still \theirs \cite{tandon-etal-2020-dataset} which we build on.

\paragraph{Entity Salience}
A small body of work on entity salience has focused on annotating entity salience in news articles and web pages for better information retrieval, recommendation, and linking \citet{gamon2013identifying,dunietz-gillick-2014-new,dojchinovski-etal-2016-crowdsourced,trani2018sel,wu-etal-2020-wn}. In contrast, we focus on entities in procedural texts, situating our work in script learning, robotic execution, automatic planning and reasoning, etc. Due to this mismatch of purpose, the definition, annotation process, and downstream applications of our entity salience and theirs are all fundamentally different.

\paragraph{Procedures and Scripts}
Script learning \cite{schank_scripts_1977} is an umbrella discipline that focuses on groups of human actions under certain scenarios. Regarding domain, procedural texts are an attractive data source to reason about entities which undergo frequent changes. There has been steady efforts in computer vision \cite{miech2019howto100m}, robotics \cite{brohan2023can}, and language \cite{9070972,https://doi.org/10.48550/arxiv.2205.07455}. In NLP specifically, work on procedures includes extracting them from instructional texts \cite{10.1145/584955.584977,delpech-saint-dizier-2008-investigating, zhang-etal-2012-automatically}, reasoning about relations among events \cite{takechi-etal-2003-feature,tandon-etal-2019-wiqa,rajagopal-etal-2020-ask,zhang-etal-2020-reasoning,https://doi.org/10.48550/arxiv.2302.13048}, knowledge-base construction \cite{jung2010automatic, chu2017distilling,park2018learning,zhou-etal-2022-show}, or applying them to downstream applications \cite{yang-etal-2021-visual, yang2021induce,zhang-etal-2020-analogous,lyu-etal-2021-goal,dalvi-etal-2019-everything,zhang-etal-2020-intent, chen-etal-2020-hybridqa}. As discussed in many of these cited works, knowledge acquired from learning scripts and procedures has been known to benefit robotics and planning.

\section{Conclusion}
We propose \ours, an improved dataset on open-domain entity tracking in procedural texts. \ours features canonicalization of entities and attributes, based on which we perform a comprehensive benchmarking evaluation of state-of-the-art LMs. \ours also provides human annotation, model prediction, and analyses of entity salience, using which we show qualitative examples on its effective on various downstream tasks. 

\section*{Limitations}
\ours, just like its predecessor \theirs, includes procedures from wikiHow which may result in homogeneous domains, writing styles, and potentially though unlikely biased, erroneous or unsafe information. Regarding canonicalization, due to the limitation of models and the imperfect human annotations in \theirs, there still exists false negatives while evaluating with metrics based on exact-match. Regarding entity salience, the definition of ``how indispensable an entity is in executing the procedure'' is motivated empirically downstream tasks and may benefit from refinement or theoretical support. The evaluation could be more trustworthy given more annotators and more procedures to be annotated. The chosen downstream tasks in this work might not be representative of all use cases of entity tracking. 

\section*{Acknowledgements}
We thank Yash Kumar Lal for providing insights into model performance, and Peter Clark for help with presentation of this paper. This work is supported in part by the DARPA KAIROS Program (contract FA8750-19-2-1004), AFRL (contract FA8750-23-C-0507), the Office of the Director of National Intelligence (ODNI) via the IARPA HIATUS Program (contract 2022-22072200005), the NSF (Award 1928631), and gifts from Roblox and Salesforce. Approved for Public Release, Distribution Unlimited. The views and conclusions contained herein are those of the authors and should not be interpreted as necessarily representing the official policies, either expressed or implied, of DARPA, ODNI, IARPA, NSF, AFRL, the U.S. Government, or of Roblox or Salesforce. The U.S. Government is authorized to reproduce and distribute reprints for governmental purposes notwithstanding any copyright annotation therein.

\bibliography{anthology,custom}

\appendix

\section{Licensing}
Our proposed \ours dataset, along with the \theirs and CREPE datasets used in this work, are under MIT license. These datasets have been and should be accessed for research purposes. During the creation of \ours, we made sure that no PII is included.

\section{Prompts of Evaluation of Entity Tracking}
\label{sec:details_eval_entity_tracking}

Example prompts pertaining to \S\ref{sec:eval_entity_tracking} are shown in Table \ref{tab:schemata_prompt_davinci}, \ref{tab:schemata_prompt_chatgpt}, \ref{tab:states_prompt_davinci}, and \ref{tab:states_prompt_chatgpt}.

\begin{table*}
\small
\centering
    \begin{tabular}{l}
        \toprule
        Content \\
        \midrule
        \makecell[l]{A person's goal is to bake a cake. For each of the steps, list the involved entities and attributes THAT UNDERGO ANY\\CHANGE. For example, for the step 'heat the oven', rack (temperature) is correct, while oven (color) is wrong.\\Step: Mix the eggs with flour.\\Entities and attributes: <generation> eggs (shape), flour (color, location), mixing bowl (content, weight)}  \\
        \bottomrule
    \end{tabular}
    \caption{Our prompt for \texttt{text-davinci-003} for the schemata prediction sub-task, followed by 1-shot demonstration.}
\label{tab:schemata_prompt_davinci}
\end{table*}

\begin{table*}
\small
\centering
    \begin{tabular}{ll}
        \toprule
        Role & Content \\
        \midrule
        User & \makecell[l]{A person's goal is to bake a cake. For each of the steps, you will list entities and attributes\\THAT UNDERGO ANY CHANGE. For example, for the step 'heat the oven', rack (temperature) is a good\\answer, while oven(color) is a bad answer. Are you ready?}  \\
        Assistant & Yes, I'm ready.\\
        User & Step: Mix the eggs with flour. \\
        Assistant & <generation>  eggs (shape), flour (color, location), mixing bowl (content, weight) \\
        \bottomrule
    \end{tabular}
    \caption{Our prompt for \texttt{gpt-3.5-turbo} for the schemata prediction sub-task, followed by 1-shot demonstration.}
\label{tab:schemata_prompt_chatgpt}
\end{table*}

\begin{table*}
\small
\centering
    \begin{tabular}{l}
        \toprule
        Content \\
        \midrule
        \makecell[l]{A person's goal is to bake a cake. For each of the steps, list all the state changes of involved entities and attributes.\\Step: Mix the eggs with flour.\\<generation for complete>  - The shape of eggs were<generation for states> solid before and fluid after.} \\
        \bottomrule
    \end{tabular}
    \caption{Our prompt for \texttt{text-davinci-003} for both the states prediction sub-task and the complete-sentence evaluation format, followed by 1-shot demonstration.}
\label{tab:states_prompt_davinci}
\end{table*}

\begin{table*}
\small
\centering
    \begin{tabular}{ll}
        \toprule
        Role & Content \\
        \midrule
        User & \makecell[l]{A person's goal is to bake a cake. For each of the steps, you will list all state changes of entities and attributes.\\You will answer in this format:\\  - attribute\_name of entity\_name was before\_state before and after\_state after\\ For example:\\  - temperature of oven was cool before and hot afterwards.\\Are you ready?}  \\
        Assistant & Yes, I'm ready.\\
        User & Step: Mix the eggs with flour. \\
        Assistant & <generation for complete>  - The shape of eggs were<generation for states> solid before and fluid after. \\
        \bottomrule
    \end{tabular}
    \caption{Our prompt for \texttt{gpt-3.5-turbo} for both the states prediction sub-task and the complete-sentence evaluation format, followed by 1-shot demonstration.}
\label{tab:states_prompt_chatgpt}
\end{table*}

\end{document}